\documentclass{llncs}
\usepackage[bottom]{footmisc}
\usepackage{enumitem}
\usepackage{todonotes}
\usepackage{gensymb}
\usepackage{booktabs}
\usepackage{tabularx}
\usepackage{amsmath,amssymb,bm}
\usepackage{multirow}
\usepackage{sidecap}
\usepackage{floatrow}
\usepackage{hyperref}

\usepackage[font=small,labelfont=bf,tableposition=top]{caption}

\DeclareCaptionLabelFormat{andtable}{#1~#2  \&  \tablename~\thetable}

\setlist[itemize]{itemsep=10pt} 
\newcommand*\samethanks[1][\value{footnote}]{\footnotemark[#1]}

\newcommand{\citet}[1]{\cite{#1}}

\begin{document}

\title{Rotation Equivariant CNNs for Digital Pathology}

\titlerunning{Rotation Equivariant CNNs for Digital Pathology}

\author{Bastiaan S. Veeling\thanks{Equal contribution.}\and Jasper Linmans\samethanks  \and Jim Winkens\samethanks 
\and Taco Cohen \and Max Welling}

\institute{University of Amsterdam, The Netherlands}

\maketitle

\begin{abstract}
We propose a new model for digital pathology segmentation, based on the observation that histopathology images are inherently symmetric under rotation and reflection.  Utilizing recent findings on rotation equivariant CNNs, the proposed model leverages these symmetries in a principled manner. We present a visual analysis showing improved stability on predictions, and demonstrate that exploiting rotation equivariance significantly improves tumor detection performance on a challenging lymph node metastases dataset. We further present a novel derived dataset to enable principled comparison of machine learning models, in combination with an initial benchmark. Through this dataset, the task of histopathology diagnosis becomes accessible as a challenging benchmark for fundamental machine learning research.
\end{abstract}

\section{Introduction}

    The field of digital pathology is developing rapidly, following recent advancements in microscopic imaging hardware that allow digitizing glass slides into whole-slide images (WSIs). This digitization has facilitated image analysis algorithms to assist and automate diagnostic tasks. A proven approach is to use convolutional neural networks (CNNs), a type of deep learning model, trained on patches extracted from whole-slide images. The aggregate of these patch-based predictions serves as a slide-level representation used by models to identify metastases, stage cancer or diagnose complications. This approach has been shown to outperform pathologists in a variety of tasks\cite{Liu2017-jq,Litjens2017-zt,Bejnordi2016-fj}.
    
    This performance is achieved using off-the-shelf CNN architectures originally designed for natural images \cite{Litjens2017-zt}. The effectiveness of these models can be largely attributed to the efficient sharing of parameters in convolutional layers. As a result, local patterns are encoded independently of their spatial location, and shifting the input leads to a predictable shift in the output. This property, known as translational equivariance, effectively exploits the translational symmetry inherent in natural images leading to strong generalization.

    In contrast to natural images, WSIs exhibit not only translational symmetry but rotation and reflection symmetry as well. CNNs do not exploit these symmetries, and as a result are found empirically to spend a large part of their parameter budget on multiple rotated and reflected copies of filters \cite{zeiler2014visualizing}.  Additionally, we find that CNNs trained on histopathology data exhibit erratic fluctuations in predictions under input rotation and reflection. Enforcing equivariance in the model under these transformations is expected to reduce such instabilities, and lower the risk of overfitting by improving parameter sharing.

    To encode these symmetries, we leverage recent findings in rotation equivariant CNNs \cite{Cohen2016-do,Worrall2017-ji,Weiler2017-oz}, a current topic of interest in the machine learning community. These methods show strong generalization under limited dataset size and are more robust under adversarial perturbations in rotation, translation and local geometric distortions \cite{Dumont2018-fa}. We propose a fully-convolutional patch-classification model that is equivariant to 90$\degree$ rotations and reflection, using the method proposed by \cite{Cohen2016-do}. We evaluate the model on the Camelyon16 benchmark \cite{Ehteshami_Bejnordi2017-pt}, showing significant improvement over a comparable CNN on slide level classification and tumor localization tasks. 
    
    As slide-level metrics potentially obscure the relative performance of patch-level models, we further validate on a patch-level task. In this regime, there is currently no benchmark that harbors the high volume, quality and variety of Camelyon16. Thus, we present \textit{PatchCamelyon}(PCam), a large-scale patch-level dataset derived from Camelyon16 data. Through this dataset, we demonstrate that the proposed model is more accurate and more robust under input rotation and reflection, compared to an equivalent standard CNN.
    
    The contributions of this work are as follows: \textbf{(1)} we propose a novel deep learning model that utilizes symmetries inherent to histopathology\footnote{\label{footnote}PCam details and data at \url{https://github.com/basveeling/pcam}. Implementations of equivariant layers available at \url{https://github.com/basveeling/keras_gcnn}.}, \textbf{(2)} demonstrate that rotation equivariance improves model reliability and \textbf{(3)} present a new large-scale histopathology dataset that enables precise model evaluation.

\par\textbf{Related Work} A common approach to improve orientation robustness is to train CNNs using extensive \textit{data augmentation}, perturbing data with random transformations \cite{Liu2017-jq,Litjens2017-zt}. Although this may improve generalization, it fails to capture local symmetries and does not guarantee equivariance at every layer. As CNNs have to learn rotation equivariance from data, they require a larger model capacity to hold copies of identical filters. Even if rotation equivariance is achieved on training data, there is no guarantee that this generalizes to a test set. Orthogonally, \citet{Liu2017-jq,Ciresan2013-wv} propose a test-time augmentation strategy that averages the predictions of 90$\degree$-rotated and mirrored versions to improve robustness to orientation-induced instability. As a downside, this comes at 8 times the computational cost and does not provide guarantees on equivariance \cite{Lenc2015-ih}.

Methods that enable equivariance under rotations and other transformations include Harmonic Networks \cite{Worrall2017-ji}, which constrain the set of filters to circular harmonics, allowing for full $360\degree$-equivariance. \cite{Weiler2017-oz} employs steerable filters and evenly samples a small number of rotations. In this work, we focus on the straight-forward G-CNN method from \citet{Cohen2016-do} applied on discrete rotation/reflection groups. Although these groups do not cover the full continuous rotational symmetry inherent in WSIs, the empirical evidence gathered so far shows that $ 90\degree$ rotation equivariance improves performance significantly\cite{Weiler2017-oz}. 

\section{Methods}
\subsection{Background}
\begin{figure}[t]
\centerline{\includegraphics[width=1.0\textwidth]{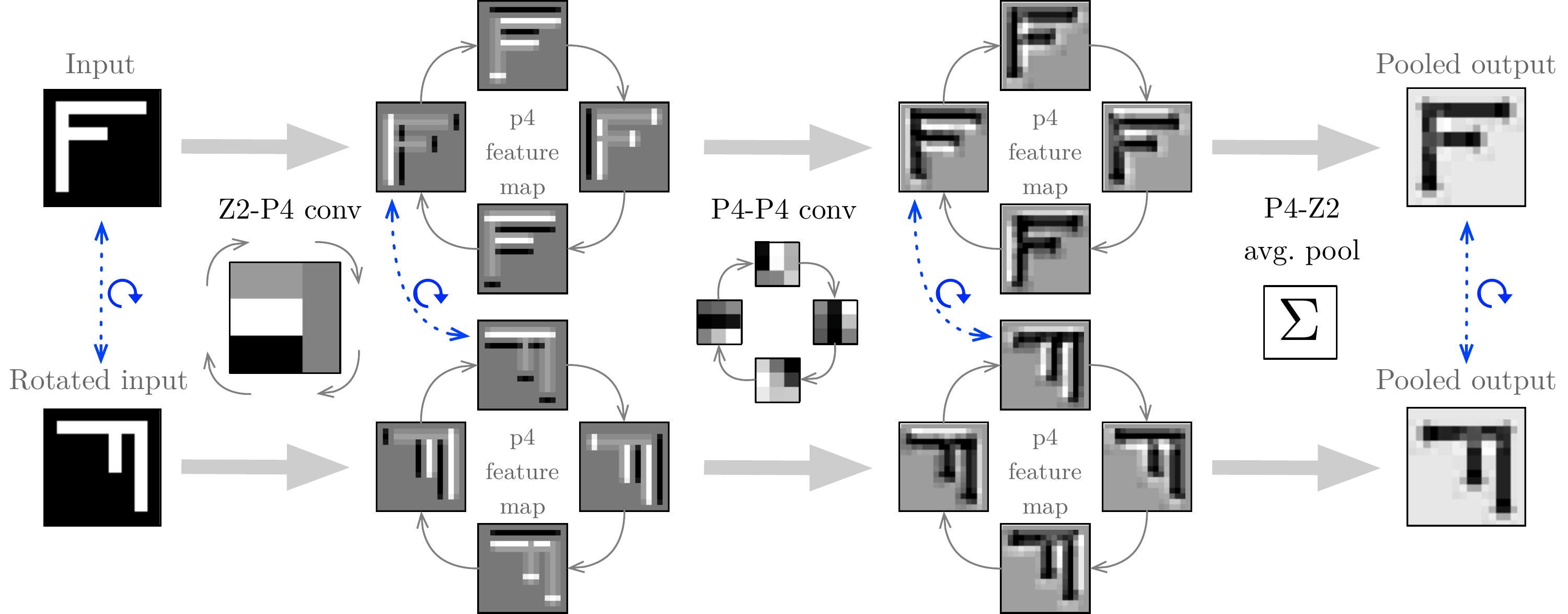}}
\caption{Given a canonical input and a rotated duplicate, we demonstrate how a 2-layer G-CNN is equivariant in $p4$. Feature maps of one kernel per layer are shown, and the dashed blue arrows indicate how (intermediate) representations of the two inputs correspond. The $\mathbb{Z}^2\rightarrow p4$ convolution correlates the input with 4 rotated versions of the same kernel. The $p4\rightarrow p4$ convolution correlates the resulting feature map with the $p4$-kernel, cyclically-shifting and rotating the kernel for each orientation. The final layer demonstrates how average-pooling over the orientations produces a representation that is locally invariant and globally equivariant to rotation. \textit{Global} average pooling over $p4$ would result in a representation globally invariant to both translation and rotation. }\label{fig:patho_equiv}

\end{figure}
In the mathematical model of CNNs and G-CNNs introduced in \cite{Cohen2016-do}, input images and output segmentation masks are considered to be functions $f : \mathbb{Z}^2 \rightarrow \mathbb{R}^K$, where $K$ denotes the number of channels, and $f$ is implicitly assumed to be zero outside of some rectangular domain.

A standard convolution\footnote{Technically, this is a cross-correlation} (denoted $*$) of an input $f$ with filter $\psi$ is defined as:
\begin{align}
    [f * \psi](x) = \sum_{y \in \mathbb{Z}^2}\sum^K_{k=1} f_k(y)\psi_k(x-y).
\end{align}

G-CNNs are a generalization of CNNs that are equivariant under more general symmetry groups, such as the group $G=p4$ of $90\degree$ rotations, or $G=p4m$ which additionally includes reflection.
In a G-CNN, the feature maps are thought of as functions on this group.
For $p4$ and $p4m$, this simply means that feature channels come in groups of $4$ or $8$, corresponding to the $4$ pure rotations in $p4$ or the $8$ roto-reflections in $p4m$.
In the first layer, these are produced using the $(\mathbb{Z}^2 \rightarrow G)$-convolution:
\begin{equation}
    [f * \psi](g) = \sum_{y \in \mathbb{Z}^2} \sum_{k=1}^K f_k(y) \psi_k(g^{-1} y),
\end{equation}
where $g = (r, t)$ is a roto-translation (in case $G=p4$) or roto-reflection-translation (in case $G=p4m$).

In further layers, both feature maps and filters are functions on $G$, and these are combined using the $(G\rightarrow G)$-convolution:
\begin{align}
    [f * \psi](g) = \sum_{h\in G}\sum^K_{k=1} f_k(h)\psi_k(g^{-1} h).
\end{align}

In the final layer, a group-pooling layer is used to ensure that the output is either invariant (for classification tasks) or equivariant as a function on the plane (for segmentation tasks, where the output is supposed to transform together with the input). In Fig.~\ref{fig:patho_equiv} we demonstrate how equivariance is achieved through this process. Non-linear activations and pooling operations are equivariant in $p4m$\cite{Cohen2016-do}, allowing layers to be freely stacked to enable deep architectures.

\subsection{G-CNN DenseNet architecture}
\begin{figure}[t]
\centerline{\includegraphics[width=1.0\textwidth]{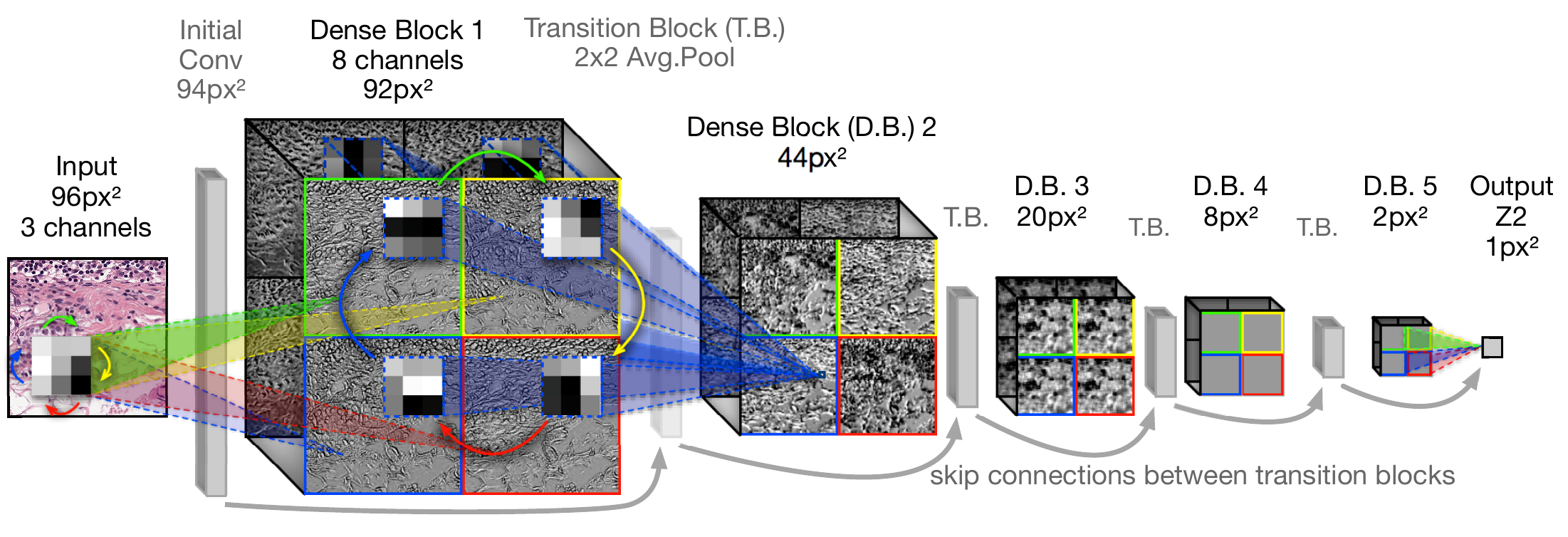}}

\caption{The proposed equivariant DenseNet architecture for the $p4$ group, consisting of 5 Dense Blocks (D.B.) alternated with Transition Blocks (T.B.). The final layer of the model is a $p4\rightarrow\mathbb{Z}^2$ group pooling layer followed by a sigmoid activation. The four orientations in $p4$ are illustrated through primary colors. A $\mathbb{Z}^2\rightarrow p4$ kernel (\textit{left}), $p4\rightarrow p4$ kernel (\textit{middle}) and $p4 \rightarrow \mathbb{Z}^2$ kernel (\textit{right}) illustrate how equivariance arises in the model.}\label{fig:model}
\end{figure}
The proposed patch-classification model is shown in Fig. \ref{fig:model} for $p4$ (the $p4m$-variant is a trivial extension). The architecture is based on the densely connected convolutional network (DenseNet) \cite{Huang2016-dq}, which consist of dense blocks with layers that use the stack of all previous layers as input, alternated with transition blocks consisting of a $1 \times 1$ convolutional layer and $2 \times 2$ strided average pooling. We use one layer per dense block due to the limited receptive field of the model, with 5 dense-block/transition-block pairs. The model spatially-pools the input by a factor of $2^5$, the output of which resembles the segmentation resolution used in \cite{Liu2017-jq}. 

Full-model group equivariance is achieved by replacing all convolution layers with group-equivariant versions \cite{Cohen2016-do}. Batch normalization layers\cite{Ioffe2015-ib} are made group-equivariant by aggregating moments per \textit{group} feature map rather than spatial feature map (as proposed by \cite{Cohen2016-do}). Zero-padding is removed to prevent boundary-effects. The final layer consists of a group-pooling layer followed by a sigmoid activation, resulting in tumor-probability output on the plane $\mathbb{Z}^2$. As the model is fully convolutional, efficient inference can be achieved at test time by reusing computation of neighbouring patches, reducing segmentation time of a full WSI from hours to $\sim2$ minutes on a NVIDIA Titan XP.

\section{Experimental results}
\subsection{Datasets and Evaluation}\label{ssec:data_eval}
To evaluate the proposed model, we use Camelyon16 \cite{Ehteshami_Bejnordi2017-pt} and PCam. Additional testing is performed on BreakHis \cite{Spanhol2016-dm}. \textbf{(1)} The Camelyon16  dataset contains 400 H\&E stained WSIs of sentinel lymph node sections split into 270 slides with pixel-level annotations for training and 130  unlabeled slides for testing.  The slides were acquired and digitized at 2 different centers  using a $40\times$ objective (resultant pixel resolution of 0.243 microns). In the Camelyon16 challenge, model performance is evaluated using the FROC curve for tumor localization. \textbf{(2)}  The PCam dataset contains 327,680 patches extracted from Camelyon16 at a size of $96\times96$ pixels @ $10\times$ magnification, with a 75/12.5/12.5\% train/validate/test split, selected using a hard-negative mining regime\textsuperscript{\ref{footnote}}.
\textbf{(3)} The BreakHis dataset contains 7909 H\&E stained microscopy images at a size of $700 \times 460$ pixels.  The task is to classify the images into benign or malignant cases for multiple magnification factors. We limit our evaluation to the images at $4\times$ magnification, for which previous approaches \cite{Spanhol2016-dm,Songyang} have reported the highest accuracy.

For the evaluation on the WSI-level Camelyon16 benchmarks, we largely follow the pipeline proposed in \citet{Liu2017-jq}, uniformly sampling WSIs and drawing tumor/non-tumor patches with equal probability. To prevent overrepresentation of background  and non-tissue patches, slides are converted to HSV, blurred, and rejected if the max. pixel saturation lies below 0.07 (range [0,1]) and value above 0.1. This was empirically verified to not drop tissue patches. For computing the FROC score, tumor location candidates are selected with an efficient square non-maximum suppression window rather than radial. The window-size is tuned per model on the validation set. FROC score confidence bounds are computed using 2000 bootstrap samples \citet{Liu2017-jq}. Train and validation splits are created by dividing the available WSIs randomly, maintaining equal tumor/normal ratio. We focus on the WSI data at $10\times$  magnification (4 times smaller than the original dataset, at 0.972 microns per pixel) to fit the compute budget available for this work. Following \cite{Liu2017-jq}, we focus on the more-granular tumor-detection FROC metric in favor of slide-level AUC.

\textbf{Training Details:} Models are optimized using Adam\cite{Kingma2014-ll} with batch size $64$ and initial learning rate $1\mathrm{e}{-3}$ (halved after 20 epochs of no improvement in validation loss). Epochs consists of 312 batches with a batch size of 64. Validation loss is computed using 40.000 sampled patches. Weights with lowest validation loss are selected for test evaluation.

\subsection{Model reliability}
\begin{figure}[t]
\includegraphics[width=\textwidth,clip,trim={0cm .05cm 0.05 0},keepaspectratio]{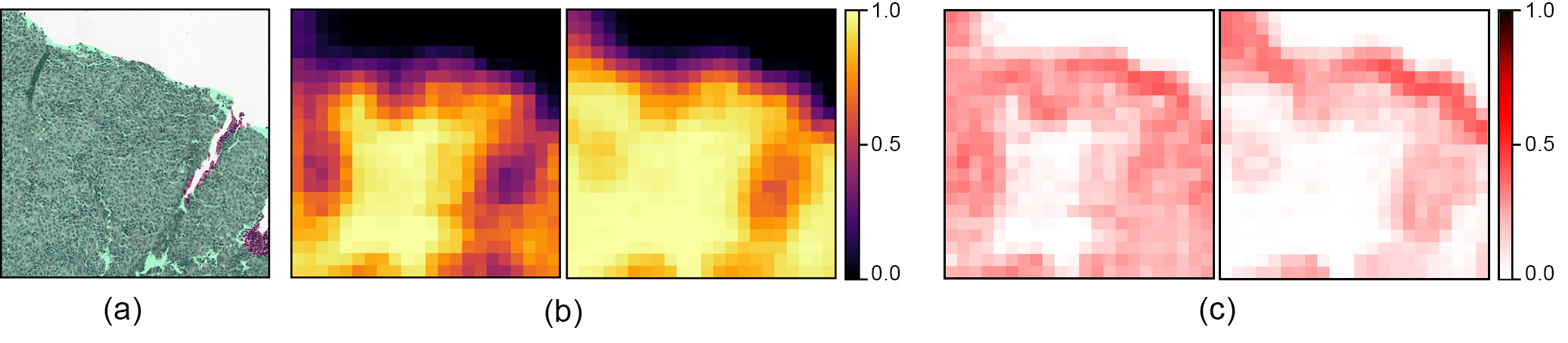}
\caption{\textbf{(a)} shows a large input region spanning multiple patches, with the tumor ground truth overlayed in green. The region is predicted under 32 evenly spaced sub-$90\degree$ rotations, and prediction maps rotated back to original orientation. \textbf{(b)} shows the mean prediction and \textbf{(c)} shows the standard deviation of the predictions across all rotations, using DenseNet (\textit{left}) and P4M-DenseNet (\textit{right}). Both networks are trained on the 12.5\% data regime. }\label{fig:rotstable}
\end{figure}

We evaluate stability of predictions under rotation of the input. We present a visual analysis in Fig. \ref{fig:rotstable}. For a comparable baseline we use an equivalent model with standard convolutions. For a fair model comparison, we keep the number of parameters consistent by multiplying the growth rate of the baseline model by the square root of the group size \cite{Cohen2016-do}. 
Bar the expected fluctuation around the tumor boundary (that arises due to the sub-sampled segmentation), the $p4m$-model is more robust to transformations even outside the group (sub-$90\degree$ rotations). In addition, we observe a higher confidence for predictions inside the tumor regions for P4M-DenseNet as compared to the baseline.

\subsection{P4M-DenseNet Performance}
\vspace{-.5cm}
\subsubsection{PatchCamelyon (PCam)}

\begin{SCtable}[]

\centering

\begin{tabular}{@{}lrrrrr@{}}
\toprule
Network & $K$ & \#$W$  &  NLL & Acc& AUC\\ 
\midrule

P4M-DenseNet   & 64 & 119K  & \textbf{0.260}  & \textbf{89.8}       &   \textbf{96.3 }  \\
P4M-DenseNet M & 24 & 19K  & 0.273   & 89.3            &   95.8  \\
P4-DenseNet    & 48 & 125K  & 0.329   & 89.0             &   94.5         \\
DenseNet+      & 24 & 128K  & 0.306    & 88.1            &   95.1          \\
DenseNet+   M  & 64 & 902K  & 0.365   & 87.2             & 94.6      \\
DenseNet       & 24 & 128K  & 0.315   & 87.6             &   95.5          \\
 \bottomrule
 
\end{tabular}

\caption{Performance on PCam, measured by negative log-likelihood, accuracy and AUC. Experiments with additional data augmentation with 90$\degree$ rotations and reflections are marked by $+$. \textit{M} indicates matching number of $\mathbb{Z}^2$ maps, \#$W$ number of weights, $K$ number of $\mathbb{Z}^2$ maps per layer.}

\label{tab:patch}

\end{SCtable}

We assess the performance of our main contribution, the P4M-DenseNet architecture, on the PCam dataset. Table \ref{tab:patch} reports the performance. P4M-DenseNet outperforms other models, closely followed by the P4-DenseNet, indicating that both rotation and reflection are useful inductive biases, that can not be learned by data augmentation alone. Keeping the number of $\mathbb{Z}^2$ maps fixed in the baseline degrades performance further, demonstrating the sample-efficiency of the P4M model.

\subsubsection{Camelyon16}
\begin{figure}[t]

\begin{minipage}[b]{0.48\textwidth}
 \begin{tabular}{@{}lll}
\toprule
Model                   & Data  & FROC                     \\ \midrule
\multirow{4}{*}{\shortstack[l]{P4M-DenseNet \\ 123k params}} & $100\%$              
                                               & 84.0	(75.5, 91.5)      \\
                        & $50\%$               & 81.5	(72.2, 89.3)                \\
                        & $25\%$               & 72.6	(58.7, 84.6)         \\
                        & $12.5\%$             & 60.7	(46.0, 74.1)                \\ \midrule
\multirow{4}{*}{\shortstack[l]{DenseNet \\ 126k params}}    & $100\%$              & 81.7 (72.1, 90.3)                                        \\
                        & $50\%$               &   80.0 (69.3, 89.1)                     \\
                        & $25\%$               &    71.0 (57.7, 82.0)                                   \\
                        & $12.5\%$             &     55.4 (42.6, 68.5)                 \\ \midrule
Liu \textit{et al.} \cite{Liu2017-jq} $10\times$               &        $100\%$          & 79.3 (74.2, 84.1) \\

Wang \textit{et al.} \cite{Wang2016-yf}              &        $100\%$          &80.7*  \\

Pathologist   \cite{Ehteshami_Bejnordi2017-pt}       &     --             &    73.3                 \\ \bottomrule
\end{tabular}
\captionlistentry[table]{A table beside a figure}
    \captionsetup{labelformat=andtable}
    \caption{Performance on the Camelyon16 test set. The confidence bounds are obtained using a 2000-fold bootstrap regime. *Challenge winner \cite{Wang2016-yf} uses 40$\times$ resolution and is not directly comparable.}
    \label{fig:cam16}
\end{minipage}
\begin{minipage}[b]{0.44\textwidth}

\includegraphics[width=\textwidth,trim={0.1cm .3cm .2cm 0},clip]{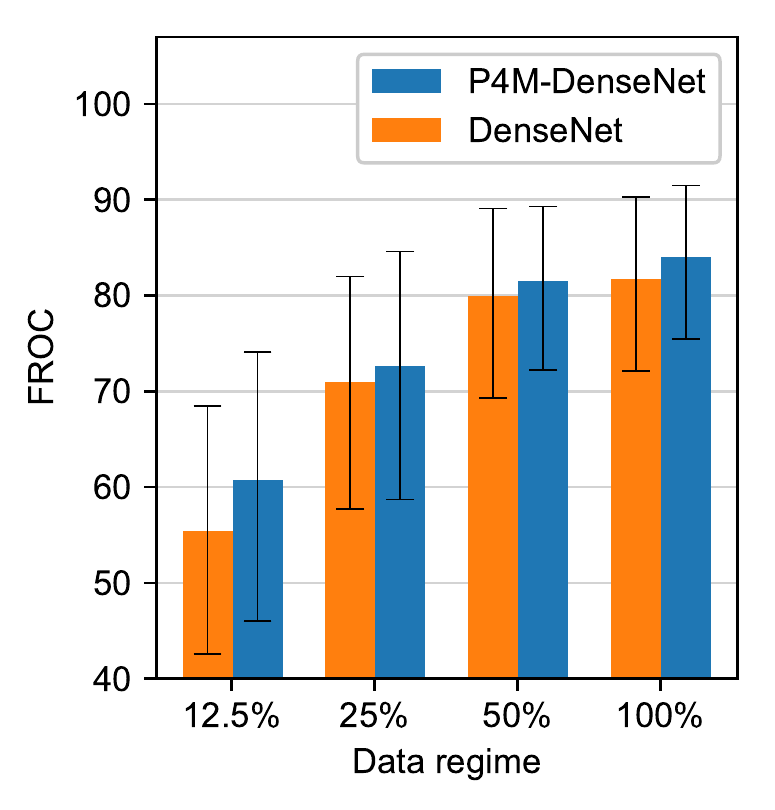}
\end{minipage}
\end{figure}

We evaluate our patch-based model on the slide-level tumor localization task of the Camelyon16 challenge. Fig. \ref{fig:cam16} reports the performance on the FROC score, next to those of a pathologist \cite{Ehteshami_Bejnordi2017-pt} and the state-of-the-art approaches reported on this dataset, including \cite{Liu2017-jq,Wang2016-yf}. For the baseline DenseNet, the training data is augmented with $90\degree$ rotations and reflection. We experiment with multiple data regimes, where the number of WSIs in the training set is incrementally reduced by a factor of two. 

The results indicate that the proposed method performs consistently better than all compared methods in terms of the FROC metric. Comparing to the baseline DenseNet results, we see that the superiority of our proposed architecture is predominantly due to the increased parameter sharing by the $p4m$-equivariance, which frees up model capacity and reduces the redundancy of detecting the same histological patterns in different orientations.

We also observe that the performance gap between our model and the baseline increases when we limit the dataset size by removing WSIs. This seems to indicate that the performance in the small-data regime benefits significantly from the sample efficiency of P4M-DenseNet, with diminishing returns when the amount of data is sufficient for the baseline network to achieve (approximate) rotation equivariance. This performance gap remains for the full data set.

\subsubsection{BreakHis} 
As an additional evaluation method, we assess the performance of the proposed model on the binary classification task of BreakHis as described in Section \ref{ssec:data_eval}. As training the model from scratch is impractical given the small dataset, we pre-train on Camelyon16 at a similar pixel resolution. Similar to \cite{Spanhol2016-dm}, we predict the malignancy of a test image by using the maximum activation of 1000 random crops. We obtain an accuracy of $96.1\pm3.2$ and $93.5\pm4.7$ for P4M-Densenet and the baseline respectively, outperforming previous approaches \cite{Spanhol2016-dm}\cite{Songyang}.

\section{Conclusion}
We present a novel histopathology patch-classification model that outperforms a competitive traditional CNN by enforcing rotation and reflection equivariance. A derived patch-level dataset is presented, allowing straightforward and precise evaluation on a challenging histopathology task. We demonstrate that rotation equivariance improves reliability of the model, motivating the application and further research of rotation equivariant models in the medical image analysis domain.
\par\textbf{Acknowledgements}
We thank Geert Litjens, Jakub Tomczak, Dimitrios Mavroeidis and the anonymous reviewers especially for their insightful comments. This research was supported by Philips Research, the SURFSara Lisa cluster and the NVIDIA GPU Grant.

\bibliographystyle{splncs}
\bibliography{main}

\end{document}